\documentclass[journal,transmag]{IEEEtran}
\ifCLASSINFOpdf
\else
\fi
\usepackage{cite}
\usepackage{array}
\usepackage{algorithmic}
\usepackage{amsmath,amssymb,amsfonts}
\usepackage{algorithm}
\usepackage{graphicx}
\usepackage{epstopdf}
\usepackage{subfigure}
\usepackage{multirow}
\usepackage{times}
\usepackage{color}
\usepackage{epsfig}

\begin{document}
%
\title{SCC-rFMQ Learning in Cooperative Markov Games with Continuous Actions}

\author{\IEEEauthorblockN{Chengwei Zhang \IEEEauthorrefmark{1},
Xiaohong Li\IEEEauthorrefmark{2},
Jianye Hao\IEEEauthorrefmark{3},
Siqi Chen\IEEEauthorrefmark{4},
Karl Tuyls\IEEEauthorrefmark{5},
Zhiyong Feng\IEEEauthorrefmark{3},\\
Wanli Xue\IEEEauthorrefmark{2}, and
Rong Chen\IEEEauthorrefmark{1}}
\IEEEauthorblockA{\IEEEauthorrefmark{1}Information Science and Technology College, Dalian Maritime University, China}
\IEEEauthorblockA{\IEEEauthorrefmark{2}School of Computer Science and Technology,
Tianjin University, China}
\IEEEauthorblockA{\IEEEauthorrefmark{3}School of Computer Software, Tianjin University, China}
\IEEEauthorblockA{\IEEEauthorrefmark{4}Southwest University, China}
\IEEEauthorblockA{\IEEEauthorrefmark{5}University of Liverpool, UK}
\thanks{Corresponding author: Jianye Hao (email: jianye.hao@tju.edu.cn).}}

%



\IEEEtitleabstractindextext{%
\begin{abstract}
Although many reinforcement learning methods have been proposed for learning the optimal solutions in single-agent continuous-action domains, multiagent coordination domains with continuous actions have received relatively few investigations. In this paper, we propose an independent learner hierarchical method, named Sample Continuous Coordination with recursive Frequency Maximum Q-Value (SCC-rFMQ), which divides the cooperative problem with continuous actions into two layers. The first layer samples a finite set of actions from the continuous action spaces by a re-sampling mechanism with variable exploratory rates, and the second layer evaluates the actions in the sampled action set and updates the policy using a reinforcement learning cooperative method. By constructing cooperative mechanisms at both levels, SCC-rFMQ can handle cooperative problems in continuous action cooperative Markov games effectively. The effectiveness of SCC-rFMQ is experimentally demonstrated on two well-designed games, i.e., a continuous version of the climbing game and a cooperative version of the boat problem. Experimental results show that SCC-rFMQ outperforms other reinforcement learning algorithms.
\end{abstract}
}

\maketitle

\IEEEdisplaynontitleabstractindextext

%
\IEEEpeerreviewmaketitle

\section{Introduction}

A large number of multiagent coordination domains involve continuous action spaces, such as robot soccer ~\cite{Riedmiller2009RLR} and multiplayer online battle arena game \cite{MENG2015190Channels}. In such environments, agents not only need to coordinate with other agents towards desirable outcomes efficiently but also have to deal with infinitely large action spaces. These problems pose great challenges to effective coordination among agents.

On the one hand, various approaches have been proposed to address the coordination problems in cooperative multiagent systems (MASs). The most common algorithms are based on Q-learning, such as the hysteretic $Q$-learning \cite{matignon2007}, Lenient-learning \cite{Panait2006Lenient}, Lenient FAQ \cite{Bloembergen2011ETS} and LMRL2 \cite{Wei2016LLI}. Although these algorithms can address coordination problems in cooperative multiagent systems to some degree, such as non-stationarity and stochasticity problems \cite{Matignon2012}, all of them are designed for discrete action spaces.


On the other hand, many research efforts have been devoted to dealing with single-agent problems with continuous action spaces. The most common algorithms are based on function approximation, which can be classified into two categories: value approximation algorithms \cite{pazis2011Learning,Sutton2009FGM,Pazis2011Generalized,lillicrap2015continuous} and policy approximation algorithms \cite{Peters2008SIR,vanHasselt2012}. Value approximation algorithms approximate the value function over the combined state-action space, while policy approximation algorithms learn over the probability density function of the continuous actions directly. The learning performance of function approximation algorithms heavily relies on assumptions on the shape of the value function, which are not always satisfied in highly non-linear problems. The recent advance of deep reinforcement learning has \cite{Silver2014Deterministic,Mnih2013Playing,Mnih2015Human,lillicrap2015continuous,DBLP17} used a deep neural network to approximate the value function. Another class of algorithms is Monte-Carlo-based methods \cite{Sallans2004RLF,Lazaric2007Reinforcement} which use re-sampling strategies and leveraging existing algorithms for discrete actions. All the above works are designed to learn optimal strategy in single-agent environments and cannot be used to solve cooperative problems in multiagent systems directly.

There are also a few works dealing with problems in continuous action multiagent games, such as learning towards fairness \cite{deJong2008AAL}, and a theoretical model \cite{Galstyan2013CSR} which analyzing steady state solutions of a continuous Boltzmann exploration based strategy in multiagent setting. Those methods are not designed for coordination in cooperative domains. Recent work MADDPG \cite{NIPS2017_7217} extends the recent deep reinforcement learning algorithm called DDPG \cite{lillicrap2015continuous} into multiagent cooperate-compete environments. This approach can be viewed as incorporating joint action leaning into DDPG critic design, which, however, does not have any mechanism to handle multiagent cooperative challenges such as equilibrium-selection, non-stationarity and stochasticity problems.






In this work, we propose a reinforcement learning method for multiagent continuous-actions coordination problems named Sample Continuous Coordination with recursive Frequency Maximum Q-Value (SCC-rFMQ). SCC-rFMQ divides the problem into two layers: (1) the action set modification layer; (2) the evaluation updating layer. The first layer extracts a discrete action set from the original continuous action spaces by a variation of re-sampling mechanism inspired by SMC \cite{Lazaric2007Reinforcement}. The new re-sampling mechanism preserves good actions of each agent and uses a variable exploratory rate to control the resample distribution and the convergence of the re-sampling mechanism. During the action set modification period, the variable exploratory rate is adjusted by a strategy named the Win or Learn More (WoLM) principle, to handle the equilibrium-selection problem, non-stationarity problem and stochasticity problem in cooperative MASs. In the evaluation updating layer, we extend the rFMQ learning methods \cite{Matignon2012} to multi-state environments such that it can handle coordination problems in more complicated environments. Experimental results show that SCC-rFMQ outperforms other state-of-the art single-agent and multiagent algorithms (SMC, CALA and MADDPG) for continuous actions in terms of both learning efficiency and effectiveness. 

The rest of this paper is organized as follows. Section 2 reviews basic notations of cooperative Markov game, and introduces SMC learning and rFMQ method briefly. Section 3 presents the proposed SCC-rFMQ learning algorithm. Experimental results are presented in Section 4 and  related works are discussed in Section 5. Finally, Section 6 draws conclusions and points out future research directions.

\section{Background}
\label{background}

\subsection{Continuous Action Cooperative Markov Game}
\label{subsection2.1}
Markov game is an extension of Repeated game (repeated interaction) and Markov Decision Process (multiple states), modeled as a 5-tuple $\langle S,N,\mathbf{A}_i,T,\mathbf{R}_i\rangle$ : $S$, the set of states; $N$, the set of players; $\mathbf{A}_i$, the action space of player $i$; $T:S\times \mathbf{A}\times S\rightarrow [0,1]$, the state transition function; and $\mathbf{R}_i:S\times \mathbf{A}\rightarrow \mathbb{R}$, the payoff function of player $i$. Here $\mathbf{A}=\mathbf{A}_1 \times ... \times \mathbf{A}_N$, $\mathbf{A}_i\in [0,1]$ for any player $i$ in $N$.

The Cooperative Markov Game is a Markov game with competition between groups of players, in which a group of players work together to achieve a specific purpose. Especially, if all players in the game receive the same reward, the game is a fully cooperative Markov game. The action space of players in Markov games could be discrete or continuous. In real-world environments, people often need to complete some tasks in which high-precision control is needed, and actions slightly different from the optimal one leads to a very low utility. Therefore, it is significant to address the problem of continuous action spaces.

\begin{table}
\begin{center}
\caption{The Climbing Game} \label{tab:FCG}
\begin{tabular}{ccccc}
\hline
\rule{0pt}{10pt}
\multirow{2}{0.6 in}  &  & \multicolumn{3}{c}{Agent 2's actions} \\
\cline{2-5}
\rule{0pt}{10pt}
 & & A & B & C \\
\hline
\multirow{3}{0.5 in}{Agent \\1's\\  actions}
 & A & 11 & -30 & 0 \\
\cline{2-5}
\rule{0pt}{10pt}
 & B & -30 & 7 & 6 \\
\cline{2-5}
\rule{0pt}{10pt}
 & C & 0 & 0 & 5 \\
\hline
	\end{tabular}
\end{center}
\end{table}

\begin{table}
\begin{center}
\caption{The Partially Stochastic Climbing Game} \label{tab:PSCG}
\begin{tabular}{cccccc}
\hline
\rule{0pt}{10pt}
\multirow{2}{0.6 in}  &  & \multicolumn{3}{c}{Agent  2's actions} \\
\cline{2-5}
\rule{0pt}{10pt}
 & & A & B & C \\
\hline
\multirow{3}{0.5 in}{Agent \\ 1's \\ actions}
 & A & 11 & -30 & 0 \\
\cline{2-5}
\rule{0pt}{10pt}
 & B & -30 & 14/0 & 6 \\
\cline{2-5}
\rule{0pt}{10pt}
 & C & 0 & 0 & 5 \\
\hline
	\end{tabular}
\end{center}
\end{table}
In a cooperative Markov game, there are many factors that may cause an algorithm fails to converge to a good solution, such as the equilibrium-selection problem, non-stationarity problem and stochasticity problem. Here we introduce the Climbing Game (CG) and the Partially Stochastic Climbing Game (PSCG) \cite{Matignon2012}, to explain these problems. The CG game in Table \ref{tab:FCG} is a fully cooperative matrix game. Each agent makes decision among three actions $A$, $B$ and $C$. Both agents receive the same payoff in the matrix corresponding to their joint action. PSCG (Table \ref{tab:PSCG}) is an extension of CG. The game is the same as CG except that the reward of joint action $\langle B,B\rangle$ is $14$ or $0$ with equal probability. Both games have two Nash equilibria, i.e., $\langle A,A\rangle$ and $\langle B,B\rangle$. Meanwhile, $\langle A,A\rangle$ is the Pareto-dominate optimal equilibrium.

The best strategy of each agent in CG game depending on the other agent's strategy. Specifically, if initial strategies of an agent is to select an action with equal probability, then action $C$ will be the best choice for the other agent because the expected return of each actions satisfies $E[v(a)]=(11-30+0)/3<E[v(b)]=(-30+7+6)/3<E[v(c)]=(0+0+5)/3$. Further, if an agent chooses $C$ with higher probability, then the other agent's best strategy will be action $B$, which means that algorithms based on the above assumptions may converge to $\langle B,B\rangle$ with high probability than $\langle A,A\rangle$. How to design an algorithm that can converge to $\langle A,A\rangle$ instead of $\langle B,B\rangle$ in the situation that sub-optimum is much more easier to be explored than global-optimum, is named \emph{equilibrium-selection problem}. Besides, since the other agent is also dynamically adjusting its strategy to optimize its reward, an agent needs to find the optimal strategy in a situation where markov property are no longer satisfied, which are critical for learning algorithms to guarantee the convergence in single agent environments. The cooperative problem caused by not meeting markov properties is named \emph{the non-stationarity problem}. Further, when the reward function is stochastic, the noise in the environment and the behaviors of other agents may both result in the variation of the reward, which makes the source of variation difficult to distinguish, see PSCG game (Figure \ref{tab:PSCG}). The randomness of the environment further increases the difficulty of finding a best strategy for a learning algorithm. This problem is named as \emph{the stochasticity problem}. Algorithms which did not consider these problems may fail to converge to $\langle A,A\rangle$.

\subsection{SMC-learning for Continuous Action Spaces}
\label{subsection2.2}

SMC-learning  \cite{Lazaric2007Reinforcement} is a sampling-based actor-critic approach on reinforcement learning with continuous action spaces. 
In the SMC-learning algorithm, the actor represents its evolving stochastic policy using Monte Carlo sampling. For every state $s$, a set of action samples $A(s)$ is maintained. An importance weight $\omega_i$ is associated with every action sample $a_i$. In action selection step, the actor takes one action randomly among those available actions in the current state, and the probability of extraction of each action is equal to its weight $\omega_i$. After the actor selects the action, the critic computes an approximation of the action-value function based on the rewards. Finally, the actor updates the policy distribution. The actor modifies the weights $\omega_i$ using the importance sampling principle and performs the policy improvement step based on the action values which are computed by the critic. In this way, actions with higher estimated utility would get more weight.

When the set $A(s)$ contains some samples whose estimated utilities are very low, the actor modifies the set of available actions by re-sampling new actions. All samples in a state approximate a probability density function over the continuous action space for that state. New action samples are drawn from this distribution through importance sampling. The weight of a sample is set proportional to the expected return of that action. Therefore, the approximated probability density function has high values where actions have high values of the expected return and are sampled and executed more often.

SMC-learning is designed for learning the optimal policies in MDPs with continuous action spaces and cannot be used in the multiagent learning environments directly.  In this work, we extend SMC-learning to multiagent settings by introducing two novel components (Coordination Resample (CR) Strategy and Multi-state rFMQ Strateg) specifically designed for addressing multiagent coordination problems.

\subsection{rFMQ algorithm for Cooperative Games }
\label{subsection2.3}
The recursive FMQ \cite{Matignon2012} is a modification of $Q$-learning to addressing the coordination problem in the cooperative game. To solve the equilibrium-selection problem and the non-stationarity problem, rFMQ maintains both the ordinary $Q$-value and the maximum reward value $Q_{max}(a)$, together with a frequency value $F(a)$ which is an estimation of the frequency of receiving the maximum reward value when the agent plays action $a$. Here we call it \emph{the maximum priority principle}. For the stochasticity problem, rFMQ recursively compute the frequency using a learning rate $\alpha_f$, and we call it \emph{the recursive reduction principle}. Formally, $F(a)$ is calculated following
\[{F}(a) \leftarrow \left\{ {\begin{array}{*{20}{l}}
1&{r > Q_{max}(a)}\\
{(1 - {\alpha _f}){F}(a) + {\alpha_f}}&{r = Q_{max}(a)}\\
{(1 - {\alpha _f}){F}(a)}&{r < Q_{max}(a)}
\end{array}} \right.\]
where $r$ is the reward in the current round. The key idea of rFMQ is to evaluate the actions using linear interpolation based on the occurrence frequency $F(a)$. The evaluation value $E(a)$ of the action $a$ is updated by the following equation: $E(a)=(1-F(a))Q(a)+F(a)Q_{max}(a)$. Finally, action of each round is selected based on the evaluation value $E(a)$ following the $\epsilon$-greedy strategy: selecting an action randomly with probability $\epsilon$ and selecting the action which maximum $E$-value with probability $1-\epsilon$.

Matignon \emph{et al.} \cite{Matignon2012} has shown experimentally that the rFMQ algorithm can deal with coordination problems in partially stochastic matrix games. Here we extend it to handle coordination problems in cooperative Markov game.

\section{Sample Continuous Coordination rFMQ}
Drawing on the idea of sampling in SMC-learning and coordination method in rFMQ, we propose SCC-rFMQ (Sample Continuous Coordination with recursive Frequency Maximum Q-Value) to handle coordination problems in cooperative Markov games with continuous action spaces. Note that SCC-rFMQ is not just a simple combination of these two algorithms. The SCC-rFMQ divides the learning process into two layers: the action set modification layer, and the evaluation and policy updating layer. To handle cooperative problems, we add a cooperative strategy for each layer. In the action set modification layer, inspired by the maximum priority principle and the recursive reduction principle in rFMQ, we put forward a new resample strategy named Coordination Resample to extract the action from the continuous action spaces. In the evaluation updating layer, we extended the the rFMQ learning algorithm to multi-state environments. Details of SCC-rFMQ is shown in Algorithm \ref{alg:SCC-rFMQ}.

\begin{algorithm}[h]
\caption{SCC-rFMQ for agent $i$ with $n$ samples}
\label{alg:SCC-rFMQ}
\begin{algorithmic}[1]
\STATE For all state $s \in S$, initialize the available action set $A_{i}(s)$ by drawing $n$ samples from $\mathbf{A}_i(s)$
\STATE \textbf{for} all $s \in S$ and $a\in A_{i}(s)$ \textbf{do} \\
         ~~~~Initialize $Q_{i}(s,a)$, $Q_{i}^{max}(s,a)$,$V_i(s)$, $E_i(s,a)$ to 0,\\
         ~~~~~~ $F_i(s,a)$ to $1$, $\epsilon_i^{re}(s)$ to $1$ and $\sigma_i(s)$ to $\sigma_0 $\\\label{initialize}
\REPEAT
\STATE $s \leftarrow $ initial state
\REPEAT
\STATE Action set modification: \\
       \textbf{if} resample condition is satisfied \textbf{then} \\
         ~~~Resample $A_{i}(s)$ using the Coordination Resample strategy (Algorithm \ref{alg:SCC}) \\ \label{SCC}
\STATE Evaluation updating:\\
       \textbf{for} all $a\in A_{i}(s)$ \textbf{do} \\
         ~~Update $Q_{i}(s,a)$ using the multi-state rFMQ learning strategy (Algorithm \ref{alg:MrFMQ}) \\ \label{rFMQ}
\STATE Update state: $s \leftarrow s'$
\UNTIL{$s$ is an absorbing state}
\UNTIL{the repeated game ends}
\end{algorithmic}
\end{algorithm}

In SCC-rFMQ, the initial sample set $A_i(s)$ is a subset of the continuous action set $\mathbf{A}_i(s)\in [0,1]$ with $|A_i(s)|=n$, where all elements $a\in A_i(s)$ are randomly selected from $\mathbf{A}_i(s)$. We initialize the $\sigma_i(s)$ to $\sigma_0$ to ensure the sampling area large enough. Other parameters are initialized following the traditional settings \cite{Matignon2012} (Line \ref{initialize}). Each round in SCC-rFMQ consists of two critical steps, i.e., Action set modification (Line \ref{SCC}) and Evaluation \& Policy updating (Line \ref{rFMQ}). First, whenever the resample condition is satisfied, the sample set $A_i(s)$ is updated by the Coordination Resample strategy. The condition we used here is quite simple: every $c$ rounds for each state. 
Next, it moves to the second step of valuation updating, which evaluates actions in $A_i(s)$ using the multi-state rFQM learning strategy (Line \ref{rFMQ}). Other lines of SCC-rFQM are similar to the traditional multi-state multi-agent reinforcement learning algorithms used in Markov games.

\subsection{Coordination Resample (CR) Strategy}

This subsection introduces the first key component of SCC-rFMQ. There are two difficulties to be addressed here. The first one is how to find a better action set than the current one, and the second is how to ensure that the algorithm can converge to the best action set eventually. To solve the first problem, inspired by the maximum priority principle in rFMQ, the CR strategy preserves the currently top $n/3$ best action and resamples $1-n/3$ new actions according to a variable probability distribution. For the second problem, a variable exploratory rate is used to control the convergence of the sampling strategy by adjusting the aforementioned probability distribution adaptively. We propose a Win or Learn More (WoLM) principle to adaptively adjust the exploratory rate, which is inspired by the recursive reduction principle in rFMQ. The Coordination Resample (CR) Strategy is detailed in Algorithm \ref{alg:SCC}.

\begin{algorithm}[h]
\caption{Coordination Resample strategy of agent $i$ in state $s$}
\label{alg:SCC}
\begin{algorithmic}[1]
\STATE Find the best action ${a_{\max }} = \mathop {\arg \max }\nolimits_{a \in {A_i}(s)} {Q _i}(s,a)$;\\
\STATE Update exploratory rate $\sigma_{i}(s)$ following the WoLM principle:\\
       \textbf{if} ${a_{\max }} \neq a_{i}(s)^{*}$ \textbf{then} $\sigma_{i}(s)\leftarrow \sigma_0$\\
       \textbf{elseif} $Q_i(s,a_{\max }) \geq V_i(s)$ \textbf{then} $\sigma_{i}(s)\leftarrow \sigma_{i}(s)\delta_d$, \\
       \textbf{else} $\sigma_{i}(s)\leftarrow \sigma_{i}(s)\delta_l$ \\ \label{exploratory}
\STATE Update $V_{i}(s)$ and $a_{i}(s)^{*}$: \\
       ~~~~~$a_{i}(s)^{*} \leftarrow a_{\max }$, $V_{i}(s) \leftarrow Q_i(s,a_{\max })$ \\ \label{updateV}
\STATE Resample $A_i(s)$:\\ \label{resample}
       (1) Remain the top $n/3$ largest actions in $A_i(s)$\\
       (2) Resample $2n/3$ new actions according to $\epsilon$ strategy:\\
       ~~~~~Choice actions with probability $1-\epsilon_{i}^{re}(s)$ by normal distribution $N(a_{\max},\sigma_{i}(s))$, or by uniform distribution $U[0,1]$ with probability $\epsilon_i^{re}(s)$\\
       (3) Update $\epsilon_{re}$: $\epsilon_i^{re}(s)=\epsilon_i^{re}(s)\delta_{\epsilon^{re}}$, ($\delta_{\epsilon^{re}}<1$).
\STATE Reset $Q_i$: $\forall a\in A_i(s)$, $Q_{i}(s,a)\leftarrow 0$\label{reinitialize}
\end{algorithmic}
\end{algorithm}

Firstly, we update the exploratory rate $\sigma_{i}(s)\in [0,\sigma_0]$ for agent $i$ at state $s$ following the WoLM principle (Line \ref{exploratory}). Intuitively, agents with WoLM principle explore a more broadly space when losing and narrow down the scope of exploration when winning, since there is a high chance that the optimal action is nearby its current sampled set. Specifically, if the current largest action is changed, then reset the exploratory rate $\sigma_{i}(s)$, else if the current average reward is no less than the accumulate maximum reward $V_i(s)$, $\sigma_{i}(s)$ is decreased to $\sigma_{i}(s)\delta_d$ ($\delta_d<1$), otherwise, it is increased to $\sigma_{i}(s)\delta_l$ ($\delta_l>1$). Similar to the recursive reduction principle in rFMQ where action evaluations fluctuate between optimistic and mean evaluations according to the stochasticity of the game, the variable exploration rate $\sigma_{i}(s)$ in CR strategy ensures that the sampling range can be adaptively changed in response to the changing environment, while also increase the probability of sampling to global optimum. Thus it can handle non-stationarity and stochasticity problems caused by cooperative game during sampling periods. Then, we update the max action $a_{i}(s)^{*}$ as well as the corresponding Q-value $V_i(s)$ in state $s$ (Line \ref{updateV}). Next, the available action set $A_i(s)$ is updated by keeping the top $n/3$ largest actions and resampling $2n/3$ new actions following certain probability distribution to substitute the rest $2n/3$ actions (Line \ref{resample}). In detail, the probability distribution is controlled by a gradually decreasing parameter $\epsilon_{i}^{re}(s) \in [0,1]$ for agent $i$ on state $s$. We sample new actions with probability $1-\epsilon_{i}^{re}(s)$ according to normal distribution $N(a_{max},\sigma_i(s))$ and probability $\epsilon_{i}^{re}(s)$ according to uniform distribution $U[0,1]$. In this way, the resampling strategy is more biased toward random sampling to ensure that the algorithm can sample good actions in the early learning stages of the algorithm. As the parameter $\epsilon_{i}^{re}(s)$ become smaller, the algorithm is more biased toward exploiting the learning experience by sampling more actions near the current optimal action. When actions in $A_i(s)$ are close to the optimal action of the whole action space $\mathbf{A}_i(s)$, the exploratory rate $\sigma_{i}(s)$ of the normal distribution will be gradually decreased to a very small value until 0. Finally, $Q_{i}(s,a)$ are re-initialized to $0$ (Line \ref{reinitialize}), to ensure that each new sampled action has enough observations to obtain a relatively correct estimate of its Q-value given that the total number of observations are limited. 

\subsection{Multi-state rFMQ Strategy}

In this subsection, we introduce another principal component: Multi-state rFMQ Strategy. We extend the rFMQ learning algorithm to multi-state multiagent games, as detailed in Algorithm \ref{alg:MrFMQ}.

\begin{algorithm}[h]
\caption{The multi-state rFMQ strategy of agent $i$ }
\label{alg:MrFMQ}
\begin{algorithmic}[1]
\STATE \textbf{if} the set $A_i(s)$ is resampled \textbf{then} \\
    ~~~~\textbf{for} all $a \in A_i(s)$ \textbf{do}\\
    ~~~~~~~ $F_i(s,a)=1$, $Q_{i}^{max}(s,a)=0$, $E_i(s,a)=0$ \\
\STATE Select an action $a$ according to $\epsilon$-greedy strategy and receive reward $r$ \label{choiceAction}
\STATE update $Q$:\\
    ${Q_i}(s,a)\leftarrow(1-\alpha)Q_i(s,a)+\alpha({r+\gamma\mathop{\max}\limits_{a'}{Q_i}(s',a')})$\\ \label{updateQ}
\STATE Calculate $E_{i}(s,a)$:\\
    \textbf{if} ${r + \gamma \mathop {\max }\limits_{a'} {Q_i}(s',a')} > Q_{i}^{max}(s,a)$ \textbf{then} \\
    ~~~$Q_{i}^{max}(s,a) \leftarrow {r + \gamma \mathop {\max }\limits_{a'} {Q_i}(s',a')} $, $F_i(s,a)\leftarrow 1$ \\
    \textbf{else if} ${r + \gamma \mathop {\max }\limits_{a'} {Q_i}(s',a')} = Q_{i}^{max}(s,a)$ \textbf{then} \\
    ~~~$F_i(s,a)\leftarrow (1-\alpha_F)F_i(s,a)+\alpha_F$ \\
    \textbf{else} $F_i(s,a)\leftarrow (1-\alpha_F)F_i(s,a)$.\\
    $E_i(s,a)\leftarrow(1-F_i(s,a))Q_i(s,a)+F_i(s,a)Q_{i}^{max}(s,a)$  \label{originalrFMQ}
\end{algorithmic}
\end{algorithm}

Initially, if the set $A_i(s)$ has been resampled, $F_i(s,a)$, $Q_{i}^{max}(s,a)$ and $E_i(s,a)$ are initialized for all actions in $A_i(s)$ (Line 1). After initialization, agent $i$ selects an action $a$ based on $\epsilon$-greedy strategy, i.e., with probability $1-\epsilon_{Q}$ choice the action with largest $E$-value, or with probability $\epsilon_{Q}$ choosing an action randomly (Line \ref{choiceAction}). Then, the expected reward $Q_i(s,a)$ is updated (Line \ref{updateQ}). Next, $E_i(s,a)$ is updated (Line \ref{originalrFMQ}).  Unlike rFMQ, we use $r + \gamma \mathop {\max }\nolimits_{a'} {Q_i}(s',a')$ instead of $r$ as the estimate of the current action, which is a natural extension of the original rFMQ to multiple states. 



\section{Experimental Evaluation}
\label{experiment}
In this section, we analyze the performance of SCC-rFMQ learning algorithm through comparing with other reinforcement learning approaches. Note that the majority of related works are evaluated on two-agent discrete-action benchmark games \cite{Matignon2012,Wei2016LLI}. We make extensions to continuous-action domains and conduct our experiments in both single-state and multiple-state games.

\subsection{Single State Cooperative Game}
\label{subsection4.1}

Using the bilinear interpolation techniques \cite{saharay2016}, we construct continuous action game models based on the CG game and PSCG game shown in Table  \ref{tab:FCG} and \ref{tab:PSCG}. Bilinear interpolation is an extension of linear interpolation for interpolating functions of two variables on a rectilinear 2D grid. Details are not discussed here due to space limitation.

\begin{figure}[h!]
\centering
\includegraphics[width=80mm]{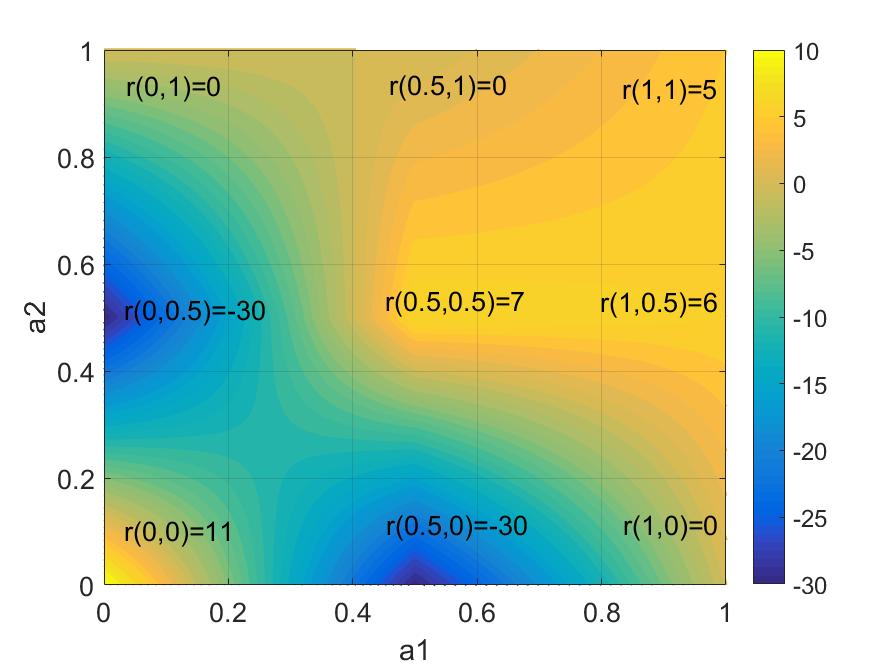}
\caption{The color map of the continuous CG}
\label{fig:colormap}
\end{figure}

To test the ability of SCC-rFMQ in handling cooperative problems in continuous action games, we replace the discrete actions with a continuous variable $a_i \in [0,1]$, where $a_i=0$, $0.5$ and $1$ stand for action $A$, $B$ and $C$ of agent $i$, respectively. The reward is defined by $r:[0,1]\times[0,1]\to R$. $r(a_1,a_2)$, $a_i \in \{0,0.5,1 \}$, satisfy rewards described in Table \ref{tab:FCG} and \ref{tab:PSCG}, while $r(a_1,a_2)$, $a_i \notin \{0,0.5,1\}$, are constructed by bilinear interpolation techniques. We use a color map (Figure \ref{fig:colormap}) to intuitively show the continuous CG.


In Figure \ref{fig:colormap}, axis $a_1$ and $a_2$ are the continuous actions of agent 1 and 2, and the rewards are represented by different colors. As it can be noticed from Figure \ref{fig:colormap}, the continuous CG has two Nash equilibriums: $\langle 0,0\rangle$ and $\langle 0.5,0.5\rangle$, and $r(0,0)=11$ is Pareto dominate than $r(0.5,0.5)=7$. Besides, we notice that the area of points whose gradients point to $\langle 0,0\rangle$ are much smaller than that of $\langle 0.5,0.5\rangle$,  and the areas of points whose average reward is higher than 7 is less than 0.001, which increase the difficulty of learning towards the Pareto-dominating optimal equilibrium $\langle 0,0\rangle$. In addition, minimal gradient nearby the sub-optimal point may easily cause agents to stabilize around local optimal.

For the continuous PSCG, we construct two deterministic continuous-action games with $r(0.5,0.5)=14$ and $0$ respectively. The reward of a joint action is obtained from the two deterministic games with equal probability.

\subsubsection{Simulation Details and Results}

In the following, we first discuss the influence of sample numbers on the performance of SCC-rFMQ, and then compare the results obtained with four different algorithms: SMC-learning \cite{Lazaric2007Reinforcement}, rFMQ \cite{Matignon2012} with different discretizations of the action space, CALA \cite{Thathachar2004Networks} and SMC+rFMQ. SMC-learning and CALA are two representative continuous-action algorithms designed for single agent games, while rFMQ is a representative multiagent coordination algorithm intended for discrete-action games. The SMC+rFMQ learning is a simple combination of the resampling strategy in SMC-learning and the multi-state rFMQ (Algorithm \ref{alg:MrFMQ}), to show the indispensable of the two basic components in SCC-rFMQ, i.e., the SCC strategy and multi-state rFMQ. For SCC-rFMQ, we use a fixed value of $c=200$, because we find that $200$ rounds are sufficient to learn a good results after extensive simulations. $\sigma_0$ is set to $1/3$ to perform a broad exploration. Details of parameter setting are listed in Appendix A.



\begin{figure}[h!]
\centering
\includegraphics[width=78mm]{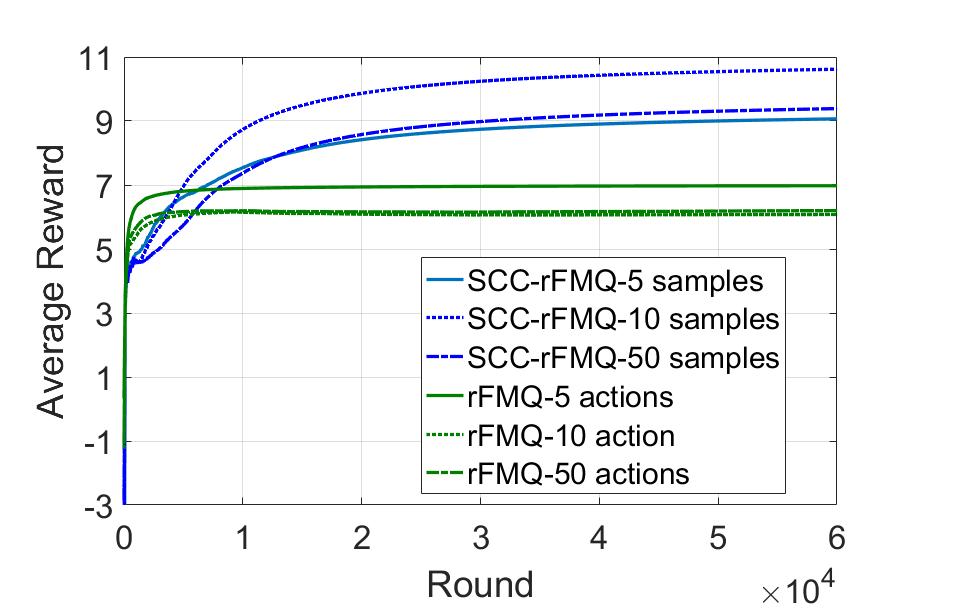}
\caption{Influences of the number of samples}
\label{fig:numberCG}
\end{figure}

\begin{figure}[h!]
\centering
\includegraphics[width=78mm]{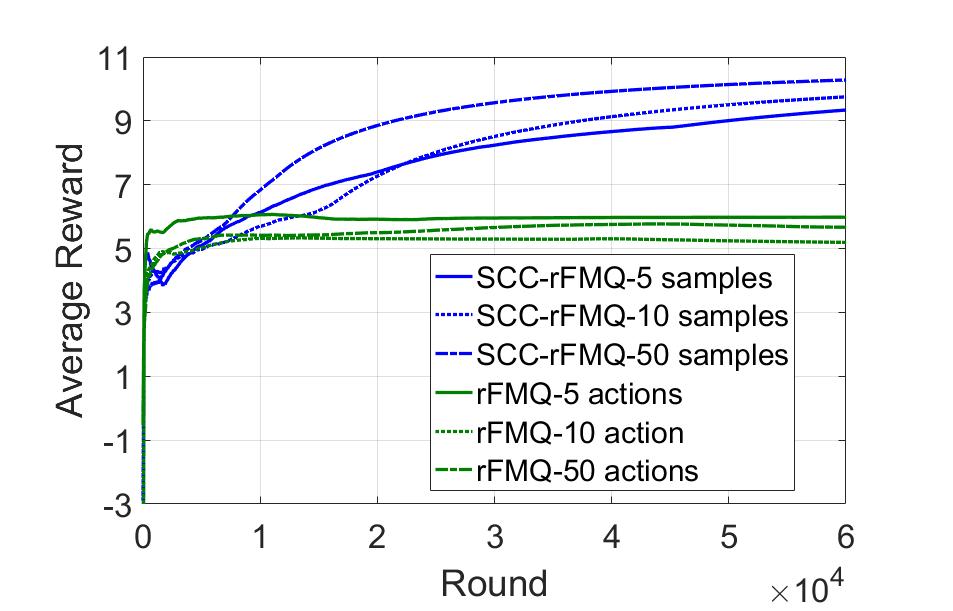}
\caption{Influences of the number of samples}
\label{fig:numberPSCG}
\end{figure}

Figure \ref{fig:numberCG} and \ref{fig:numberPSCG}  show the learning performance (in terms of cumulative average reward per episode) for SCC-rFMQ and rFMQ with 5, 10 and 50 evenly distributed actions in the continuous CG game and the continuous PSCG game respectively. For each experiment, we avoid sampling any action near the global optimal area on purpose to show the effectiveness of our algorithm. All the results are averaged over 50 runs. As it can be noticed, in all cases, our algorithm SCC-rFMQ significantly outperforms the rFMQ learning. For both game, SCC-rFMQ with different number of samples converge to more than 9, while rFMQ results converge to 7 below. Besides, the number of samples has a significant impact on the performance of SCC-rFMQ, but does not have obvious regular pattern. In the continuous CG game, SCC-rFMQ with 10 samples significantly outperforms the other two experiments, followed by SCC-rFMQ with 50 samples, and SCC-rFMQ with 5 samples performs worst, while in the continuous PSCG game, SCC-rFMQ with 50 samples performs best, followed by SCC-rFMQ with 10 samples, and finally SCC-rFMQ with 5 samples. In addition, the number of actions has little effect on the performance of rFMQ, which indicates that traditional discrete algorithms are not always suitable for continuous cooperative environments by using fine discretization.

\begin{figure}[h!]
\centering
\includegraphics[width=78mm]{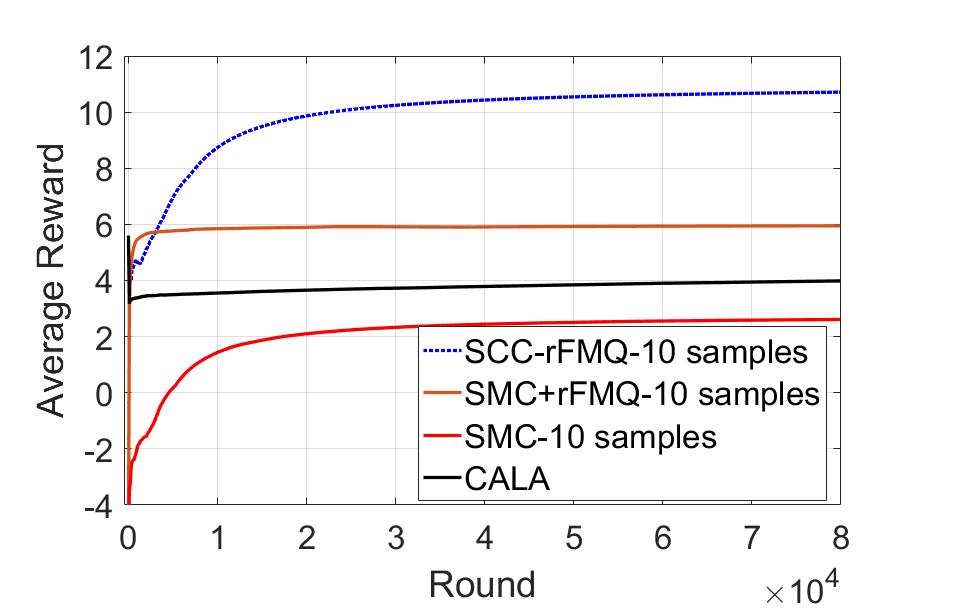}
\caption{Performance comparison between SCC-rFMQ, SMC, SMC+rFMQ and CALA in Continuous CG game}
\label{fig:performanceCG}
\end{figure}

\begin{figure}[h!]
\centering
\includegraphics[width=78mm]{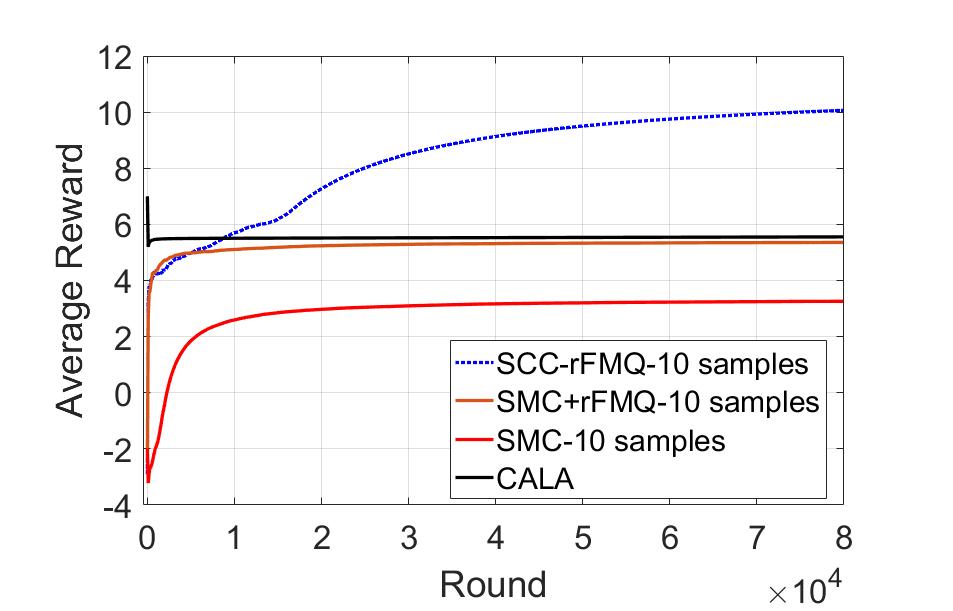}
\caption{Performance comparison between SCC-rFMQ, SMC, SMC+rFMQ and CALA in Continuous PSCG game}
\label{fig:performancePSCG}
\end{figure}
Next we compare the experimental performance of SCC-rFMQ and other algorithms. Figure \ref{fig:performanceCG} and \ref{fig:performancePSCG} show the learning performance (in terms of cumulative average reward per episode) for SCC-rFMQ, SMC, SMC+rFMQ and CALA in the continuous CG game and PSCG game respectively. Experiment settings are the same as experiments in Figure \ref{fig:numberCG} and \ref{fig:numberPSCG}. Similarly, all the results are averaged over 50 runs. For SCC-rFMQ, SMC and SMC+rFMQ, we take only 10 evenly distributed actions as the initial sample set to get a relatively fair comparison. From the figure we can see, in both games, our SCC-rFMQ significantly outperforms the other three algorithms, followed by SMC+rFMQ and CALA, and SMC-learning performs worst. It is worth mentioning that although the performance of SMC+rFMQ has a significant improvement than SMC, it is still far worse than SCC-rFMQ, which indicate that SCC-rFMQ is not a simple combination of the SMC-learning and the multi-state rFMQ.





\subsection{Multiple State Cooperative Game}
\label{subsection4.2}

To further validate the performance of our algorithm, we consider a classic multi-state continuous action space game, i.e., the boat problem, which has been initially defined in ~\cite{Jouffe1998}. Here we redefine it in a more generalize way.

\begin{figure}[h!]
\centering
\includegraphics[width=80mm]{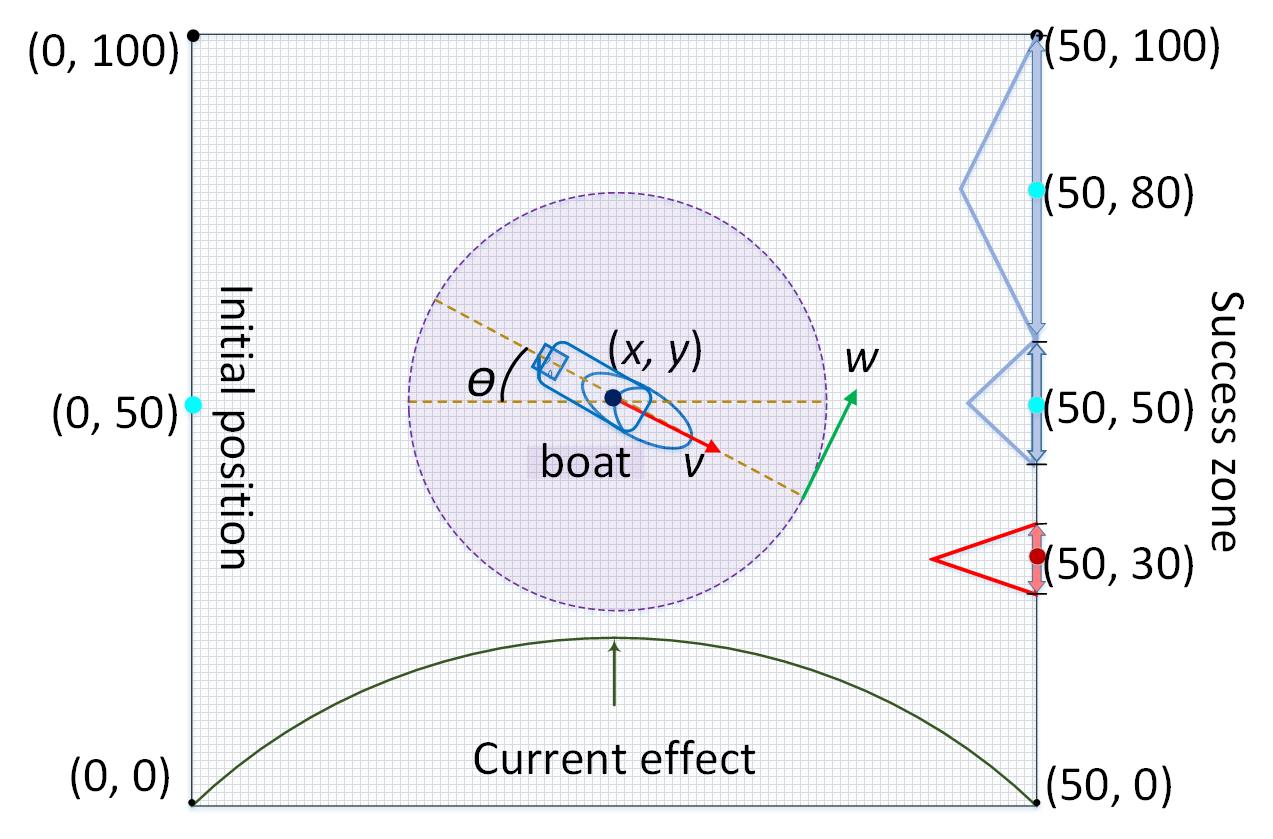}
\caption{The boat problem}
\label{fig:boat}
\end{figure}

The goal of the boat problem is to drive a boat from the quay center on the left side quay centers to quays  on the right side of the river as quickly as possible (see Figure \ref{fig:boat}). The boat is driven by two independent controllers, which control the speed and the angular speed of the boat respectively. Therefore they need to learn how to coordinate to achieve the goal. The state of the boat can be represented by a 5-tuple $\langle x,y,\theta,v,\omega\rangle$, where $x\in [0,50] $ and $y\in [0,100]$ are the boat's bow coordinates, $\theta \in [-\pi/3,\pi/3]$ is the boat's angle, $v\in [2,5]$ and $\omega\in [-1,1]$ are the boat's speed and angle speed, respectively. Ranges of above variables ensure that the boat can avoid falling into an infinite loop. The two controllers' actions are represented by two continuous variables $a_v\in [-1,2]$ and $a_{\omega} \in [-1,1]$, where $a_v $ stands for the boat's forward acceleration and $a_{\omega}$ stands for the boat's angular acceleration. The two controllers are independent of each other, which is in line with the reality. The two controllers' actions are represented by two continuous variables $a_v\in [-1,2]$ and $a_{\omega} \in [-1,1]$, where $a_v $ and $a_{\omega}$ stand for the boat's forward acceleration and angular acceleration respectively. The two controllers are independent of each other, which is in line with the reality. The quay center on the left side of the river is located at $(0,50)$. The boat's state variables are computed by the following equations:
\begin{equation}
\label{statevaluable1}
\begin{split}
\begin{array}{l}
{x_{t + 1}} = {\Pi _{[0,50]}}( {{x_t} + \int_t^{t + 1} {{v_t}\cos \left( {{\theta _t}} \right)} dt} )\\
{y_{t + 1}} = {\Pi _{[0,100]}}( {{y_t} + \int_t^{t + 1} {{v_t}\sin \left( {{\theta _t}} \right) + E( {{x_t}} )dt} } )\\
{\theta _{t + 1}} = {\Pi _{[ - \pi /3,\pi /3]}}( {{\theta _t} + \int_t^{t + 1} {{\omega _t}dt} } )\\
{v_{t + 1}} = {\Pi _{[2,5]}}\left( {{v_t} + {a_{v,t}}} \right)\\
{\omega _{t + 1}} = {\Pi _{[ - 1,1]}}\left( {{\omega _t} + {a_{\omega ,t}}} \right)
\end{array}
\end{split}
\end{equation}
where $\Pi_\Delta$ is the projection function which maps the input value to the valid range of $\Delta$ and prevents the variable moving out of the valid space. 
To increase the difficulty of the problem, we define the current effect by $E(x)={f_c}[x/50-{(x/50)}^2]$, where $f_c$ is the current force and is subject to normal distribution $N(4,1)$ at each time. This setting makes the game suffers from the stochasticity problem. Besides, we define three target quays at the right side of the river, whose quay centers are located in $(50,30)$, $(50, 50)$ and $(50,100)$ with quay widths equal to 10, 20 and 40 respectively. The immediate reward function maps a reward decreasing linearly from $0$ to $15$ for quay $(50, 30)$, or $10$ for quays $(50, 50)$ and $(50,100)$, relative to the distance from the quay center. Formally,
\[
r(x,y) = \left\{ {\begin{array}{*{20}{c}}
{15 - 3|y - 30|}&{x = 50 \wedge 25 < y \le 35}\\
{10 - |y - 50|}&{x = 50 \wedge 40 < y \le 60}\\
{10 - |y - 80|/2}&{x = 50 \wedge 60 < y \le 100}\\
0&{others}
\end{array}} \right.
\]

The overall reward of a trial is defined by $R=r(x,y)-0.1\times T, (x,y)\in Z$, where $T$ is the step number to reach the end of the trial, and $Z=\{(x,y)|x=50 \vee y=0 \vee y=100\}$. The goal of the boat game is to control the forward acceleration $a_v$ and the angular acceleration $a_\omega$ of the boat to optimize reward $R$. Intuitively, the reward of the game is determined by the running time and the final position: first, the longer of the running time, the lower the return; second, the position $(50,30)$ is the global optimal position but difficult to explore for influence of water flow and quay width, while $(50,50)$ and $(50,80)$ are suboptimal position which can be easier to explore than the global optimal position. The multiple suboptimal goal defined here can be used to test the ability of solving local optimal traps of an algorithm.

\subsubsection{Simulation Details and Results}

The goal is to train the two controllers $a_v$ and $a_{\omega}$ at each time step to coordinate the highest reward. State variables $x$ and $y$ are discredited in 1 interval. $\theta$, $v$ and $\omega$ are discredited in 10 intervals. Thus, the system has a total of 5000 thousand states. At each trial, the initial state is set to $\langle 0,50,0,2,0\rangle$. In the following, we compare our algorithm with MADDPG \cite{NIPS2017_7217} and SMC-learning \cite{Lazaric2007Reinforcement}. For each algorithm, we use two independent agents to control $a_v$ and $a_{\omega}$ separately. Beside, for SMC, we also consider another single-agent benchmark strategy which uses SMC-learning to control $a_v$ and $a_{\omega}$ simultaneously, to show the benefit of modeling it as a multiagent learning problem. MADDPG is a deep reinforcement learning method proposed to learn policies by a centralized training with decentralized execution, which has received much attention in multi-agent cooperative-competitive environments. SMC represents the class of algorithms designed for continuous-action games while having no built-in coordination mechanism. Parameters of each algorithm are set as the optimal parameters after multiple tests, and is shown in Appendix A.

\begin{figure}[h!]
\centering
\includegraphics[width=80mm]{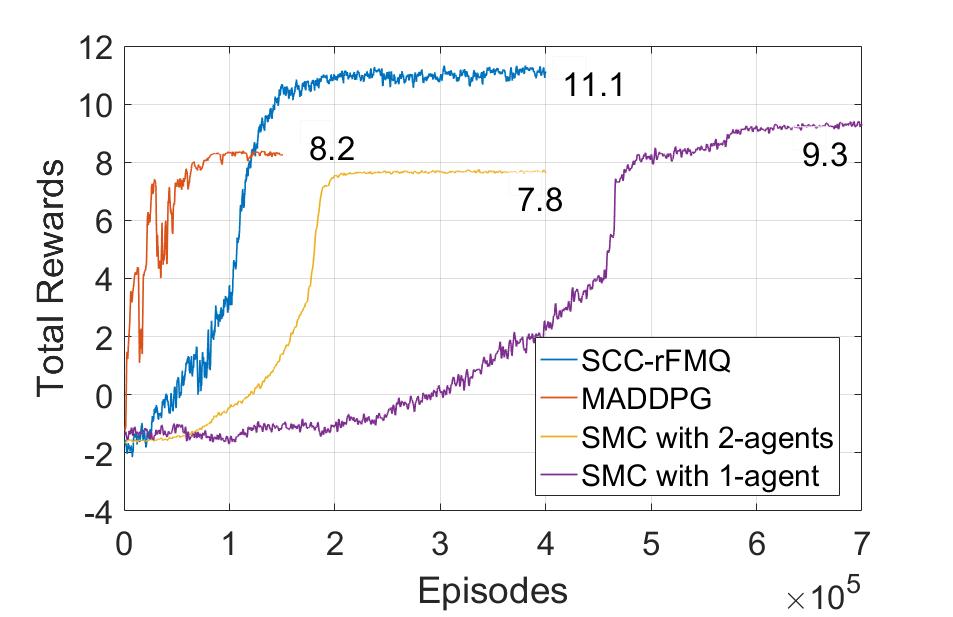}
\caption{Performance in the boat problems}
\label{fig:boatalgorithm}
\end{figure}

Figure \ref{fig:boatalgorithm} compares learning performances in terms of total reward per episode of SCC-rFMQ, SMC-learning and MADDPG averaged over 10 runs. The sample numbers of SCC-rFMQ and two-agent SMC are set to 5 and 10 respectively, and SMC in one agent experiment is set to $5 \times 5$ ($a_v$ and $a_{\omega}$ five each), which are the best settings we found during our evaluations. Besides, all initial samples of those experiments are evenly distributed from the action space. As it can be seen from Figure \ref{fig:boatalgorithm}, SCC-rFMQ outperforms SMC-learning and MADDPG regarding the final reward value, followed by the single agent SMC and MADDPG with $9.3$ and $8.2$ respectively, and SMC with multiagent learns worst. For learning speed, MADDPG converges at 100 thousands episodes, followed by SCC-rFMQ and SMC in multi-agent experiments, and lastly SMC in one-agent experiments at 600 thousands episodes. Considering that deep learning algorithms require more computing resources per episode, the result that  MADDPG takes less episodes to converges than SCC-rFMQ should not be considered as a meaningful evaluation criteria. In summary, SCC-rFMQ outperforms SMC-learning and MADDPG in the boat problem.

\subsection{Experimental Analysis}

First, the number of sample of SCC-rFMQ can affect the performance of the algorithm, for more sampling means more granular discretization of the problem and a broader exploration of the action space. On the other hand, although the increase of the number of sampling could improve agents' exploration capabilities and thus increase the chance of finding the optimal action, it would also decrease agents' ability to respond quickly to changes in the environment, for the reason that more samples means less observations per sample which affects the estimation of actions's expected return by the multi-state rFMQ.

The low performance of other algorithms in Figure \ref{fig:numberCG}, \ref{fig:numberPSCG}, \ref{fig:performanceCG} and \ref{fig:performancePSCG} are mainly caused by the minimal gradient nearby the sub-optimal points $\langle 0.5,0.5\rangle$. SMC and CALA are not designed to handle the equilibrium-selection problem, the non-stationarity problems and the stochasticity problem which are unique to multiagent scenarios. Thus they are insensitive to changes in the environment and converge in advance to where they should not be. The low performance of rFMQ is mainly caused by the sampled action set which does not contain the optimal action, since the area of points whose gradients pointing to $\langle 0,0\rangle$ are too small to be sampled. Besides, rFMQ is only suitable for games with small amount of actions, and too many actions would significantly affect its learning performance. The low performance of SMC+rFMQ is mainly due to the static resampling strategy without any built-in coordination mechanism. Another reason is that Q-values of the new sampled action set may unable to inherit from the old one in a multi-agent environment, which causes the rFMQ strategy in SMC+rFMQ failing to learn good results during the evaluation updating period.

In addition, SCC-rFMQ performs better than SMC and MADDPG in the boat problem. Low performance of MADDPG and SMC is mainly caused by the equilibrium-selection problem of this game, for the reason that the global optimal equilibrium, i.e., the Pareto-optimal Nash equilibrium of the game is much more difficult to explore than the the other two Nash equilibria. Besides, the stochasticity problem of this game, caused by the randomness of current effect (influenced by the parameter $f_c$) make algorithms even more difficult to learn the global optimal, which affects the performance of MADDPG and the two-agent SMC. Besides, it is obviously that single agent SMC outperforms SMC with multiple agent, for the reason that SMC with centralized processing does not need to consider the cooperative problems between two independent agents. The return of the single agent SMC is slightly smaller than SCC-rFMQ mainly because the global optimum is difficult to explore.


\section{Related Works}
\label{relatedworks}
\subsection{Methods for Cooperative Markov games}
The distributed Q-learning \cite{Lauer2000ICML} proposed by Lauer and Riedmiller is based on 'optimistic independent agents'. In distributed Q-learning, agents neglect the penalties and update the evaluation of an action only if the new evaluation is greater than the previous one. The algorithm can finds optimal policies in deterministic environments for cooperative Markov games. However, this approach may not converge to optimal policies in stochastic environments.

To avoid mis-coordination in stochastic games, a common strategy is varying the 'degree of optimism' of the agents, such as hysteretic Q-learning \cite{matignon2007} and Lenient-learning \cite{Panait2006Lenient,Bloembergen2011ETS}. The hysteretic Q-learning \cite{matignon2007} uses two learning rates $a$ and $b$ for the increase and the decrease of $Q$-values, where $a>b$ to obtain chiefly optimistic learners. If $Q > r + \gamma \max Q$, update $Q$ by $a$, otherwise $b$. The algorithm plays well in stochastic games, but has poor performance in robust against alter-exploration problem. Lenient-learning \cite{Panait2006Lenient} varying the 'degree of optimism' over time, in which agents are initially lenient (or optimistic) and the degree of leniency decreases as the action is often selected. Recursive FMQ\cite{Matignon2012} evaluate the actions by biasing the probability of choosing an action with the frequency of receiving the maximum reward $Qmax$ for that action, together with a linear interpolation based on the occurrence frequency and bounded by the $Qmax$ and $Q$ values. Both Lenient-learning and rFMQ perform well in partially stochastic matrix games.


\subsection{Methods for Continuous Action spaces games}

One popular class of approaches to dealing with the Continuous action problems are value-approximation algorithms, such as \cite{Sutton2009FGM,lillicrap2015continuous}. The actions and values are interpolated to form an estimate of the continuous action-value function in the current state. The main problem in these approaches is they relies on strong assumptions about the shape of the value function. Approaches of Pazis et al. \cite{Pazis2009BAS,pazis2011Learning,Pazis2011Generalized} transforms an MDP with a large number of actions to an equivalent MDP with binary actions, where action complexity is cast to state complexity, which are still belonging to value-approximation algorithms.

Policy approximation algorithms \cite{KONDA2003,Thathachar2004Networks,Peters2008SIR} circumvent the need for value functions by representing policies directly. One of their main advantages is that the approximate policy representation can often output continuous actions directly. In order to tune their policy representation, these methods use gradient descent, update the policy parameters directly. However, this class of works are sensitive to local optima.

Another approach to deal with the action selection problem is sampling\cite{Sallans2004RLF,Lazaric2007Reinforcement}. Using Monte-Carlo estimation, those algorithms are able to choose actions that have a high probability of performing well, without exhaustively searching over all possible actions. Unfortunately, the number of sample required in order to get a good estimate may be quite high, especially in large and complicated action spaces.

\section{Conclusion}

In this paper, we propose an algorithm SCC-rFMQ to address the coordination problem in the continuous action cooperative Markov games. The SCC-rFMQ includes two critical steps: Coordination Resample and multi-state rFMQ. Coordination Resample algorithm solves the continuous action spaces by resampling the available action sets, and multi-state rFMQ algorithm mainly deals with the cooperative problem in the cooperative Markov games. Simulation results reveal that our algorithm SCC-rFMQ outperforms existing single-agent and multiagent continuous-action reinforcement learning algorithms. Except for the challenge of continuous action spaces, the state space in real-world domains may also be continuous. As future work, we are planning to extend SCC-rFMQ to handle both continuous states and actions space.


%

\appendices
\section{ Parameter Setting}

Table \ref{tab-parameter} where the learning and decay rates used for the results presented in Section \ref{experiment}. Here t is the number of repetitions of the game. With the detailed algorithmic descriptions in Section 3 and these parameters detail, all of the presented results are reproducible.

\begin{table}[h!]
\caption{Learning parameters for the results presented in Section \ref{experiment}}
\label{tab-parameter}
\centering
\begin{tabular}{l|l|l}
\hline
\hline
  parameter   & description &   value    \\
\hline
\multicolumn{3}{l}{Common} \\
\hline
  $\alpha$              &  Learning rate of Q value   &   0.5                              \\
  $T$                   &  Total Learning Round       &   80000 \\
                        &                             &   (CG and PSCG )  \\
                        &                             &   800000 \\
                        &                             &   (The boat problem)             \\
  $\gamma$              &  Discount factor            &   1 (The boat problem)              \\
  $A(0)_n$              &  Initial sample action set,            &   $\{\frac{i}{n+1}| i=1,...,n\}$\\
                        &  n is the sample number                &                                  \\
\hline
\multicolumn{3}{l}{SCC-rFMQ} \\
\hline
  $c$                   &  Re-sampling condition in Alg. 1       &   $200$                               \\
  $\alpha_F$            &  Learning rate of F in Alg. 3          &   $0.01$                               \\
  $\epsilon$            &  $\epsilon$-greedy in Alg. 3           &   $\frac{10}{10+mod(t,c)}$ \\
  $\epsilon_i^{re}(s)$  &  $\epsilon$ in Alg. 2                  &   $\frac{1}{2^{\lfloor t/c\rfloor}}$ ($\delta_{\epsilon_i^{re}}=\frac{1}{2}$)  \\
  $\delta_d$            &  Reduction rate in Alg. 2              &   $0.5$              \\
  $\delta_l$            &  Growth  rate in Alg. 2                &   $1.1$              \\
  $\sigma_0$            &  Initial exploratory rate in Alg. 2    &   $0.33$             \\
\hline
\multicolumn{3}{l}{SMC} \\
\hline
  $\sigma$              &  Re-sampling condition                 &   $0.9$ (CG and PSCG)              \\
                        &                                        &   $0.8$ (two-agent SMC)             \\
                        &                                        &   $0.6$ (one-agent SMC)           \\
  $\tau $               &  Temperature                           &   $25*0.9^{\lfloor \frac{t}{5000}\rfloor}$ \\
                        &                                        &   (CG and PSCG)) \\
                        &                                        &   $10*0.9^{\lfloor \frac{t}{2000}\rfloor}$ \\
                        &                                        &   (The boat problem)\\
\hline
\multicolumn{3}{l}{rFMQ} \\
\hline
  $\epsilon$            &  $\epsilon$-greedy in rFMQ             &    $\frac{10}{10+t}$             \\
\hline
\multicolumn{3}{l}{SMC+rFMQ} \\
\hline
  $\epsilon$            &  $\epsilon$-greedy in Alg. 3           &    $\frac{10}{10+mod(t,c)}$             \\
  $\tau $               &  Temperature                           &   $20*0.9^{\lfloor t/5000\rfloor}$ \\
  $c$                   &  Re-sampling condition                 &   $200 $               \\
  $\alpha_F$            &  Learning rate of F in Alg. 3          &   $0.01$                               \\
  $w_{i}^{t+1}$         &  Policy of agent $i$                   &   $\frac{e^{\triangle Q_{i}^{t+1}(s,a)/\tau}}{\sum_{a}e^{\triangle Q_{i}^{t+1}(s,a)/\tau}}$\\
\hline
\multicolumn{3}{l}{CALA} \\
\hline
  $\lambda$        &  The learning rate                     &   $0.05$   \\
  $\sigma_L$    &  Lower bound of the variance $\sigma$  &  $10^{-5}$ \\
\hline
\multicolumn{3}{l}{MADDPG} \\
\hline
                        &  Size of $M_{tuple}$                  &  $300000$    \\
                        &  Batch size ($M_{tuple}$)              &  $1024$    \\
                        &  Update frequency                     &  once each episode \\
                        &  Maximal episode length                &  $40$       \\
                        &  Policy nets neuron number             &  $16$       \\
                        &  Output layer activation function      &  $tanh$       \\
                        &  Critic nets neuron number             &  $32$       \\
                        &  Policy learning rate                  &  $0.01$       \\
                        &  Critic learning rate                  &  $0.01$       \\
\hline
\hline
\end{tabular}
\end{table}

In Table \ref{tab-parameter}, $\epsilon$ and $\epsilon_i^{re}(s)$ are strategy valuables for SCC-rFMQ and rFMQ, which are defined as maps that decrease with increasing learning trials $t$. $A(0)_n$  defined in SCC-rFMQ, SMC and rFMQ are initial distributed action sets evenly sampled from action space $[0,1]$, where $n$ is the sampling set size.

For SMC and rFMQ, the common parameters (e.g., $\alpha_Q$ and $\gamma$) are set to be the same as SCC-rFMQ for a fair comparison. Parameter definition of $\sigma$ and $\tau$ are the same as \cite{Lazaric2007Reinforcement}, and $\lambda$ and $\sigma_L $ are the same as \cite{deJong2008AAL}. Parameter definitions of MADDPG are the same as \cite{NIPS2017_7217}, where policies and critics are parameterized by a two-layer ReLU MLP followed by a fully connected layer (activated by $tanh$ function for policy nets). For all parameters, we select the values with the best performance after extensive simulations. It should be noted that parameter changes do not significantly affect the conclusion of the experiment in our experiments. For SMC+rFMQ, we use rFMQ with $\epsilon$-greedy strategy to learn $Q$ value, and use the resampling strategy of SMC to update action set every $c=200$ episodes. Weight value $w_{i}^{t+1}(s,a)$ used in the SMC resampling strategy is calculated by the Boltzmann exploration strategy, where
\[\Delta {\rm{Q}}_i^{t + 1}(s,a) = {\rm{Q}}_i^{t + 1}(s,a) - {\rm{Q}}_i^t(s,a)\]


\section*{Acknowledgment}

The authors would like to thank...

\ifCLASSOPTIONcaptionsoff
  \newpage
\fi

\bibliographystyle{IEEEtran}
\bibliography{ijcai18}

\end{document}